\documentclass[runningheads]{llncs}
\usepackage[T1]{fontenc}
\usepackage{tabularx}
\usepackage{booktabs}   
\usepackage{caption} 
\usepackage{graphicx}
\usepackage{tikz}
\usepackage{float}
\usetikzlibrary{arrows.meta, positioning, fit, shapes.geometric, shapes.misc}
\begin{document}
\title{A Survey on Data-Driven Models for Soil Moisture Regression and Classification}

\author{Ilektra Tsimpidi\inst{1}\orcidID{0009-0002-5967-1944} \and
George Georgoulas\inst{1}\orcidID{0000-0001-9701-4203} \and
Vidya Sumathy\inst{1}\orcidID{0000-0002-5709-0591} \and
George Nikolakopoulos\inst{1}\orcidID{0000-0003-0126-1897}}
\authorrunning{I.Tsimpidi et al.}
%
\institute{Robotics and AI Team\\Department of Computer Science, Space and Electrical Engineering\\Lule\aa University of Technology\\Lule\aa, Sweden}

\maketitle              
\begin{abstract}
Soil Moisture (SM) modelling constitutes a complex spatio-temporal learning problem characterised by nonlinear environmental interactions, heterogeneous data sources, and limited ground observations. 
Physics-based approaches, such as water balance models, rely on explicit hydrological equations and high-quality inputs, but their computational cost and scalability limitations restrict large-scale deployment. 
Data-driven artificial intelligence (AI) methods have emerged as flexible alternatives, enabling the extraction of empirical relationships between soil moisture and environmental variables with reduced modelling assumptions.
This work presents a structured survey of AI-based models for soil moisture estimation and classification. 
Existing approaches are organized into five categories: (a) statistical time-series models, (b)  geostatistical methods (c)  classical machine learning (ML) models, (d) Deep Learning (DL) models and (e) Probabilistic/Bayesian methods. 
These models leverage historical soil moisture records, meteorological variables, vegetation indices, topography, soil characteristics, and geolocation data to perform regression or classification tasks.

\end{abstract}
\section{Introduction}
\begin{figure}
    \centering
    \includegraphics[width=0.6\linewidth]{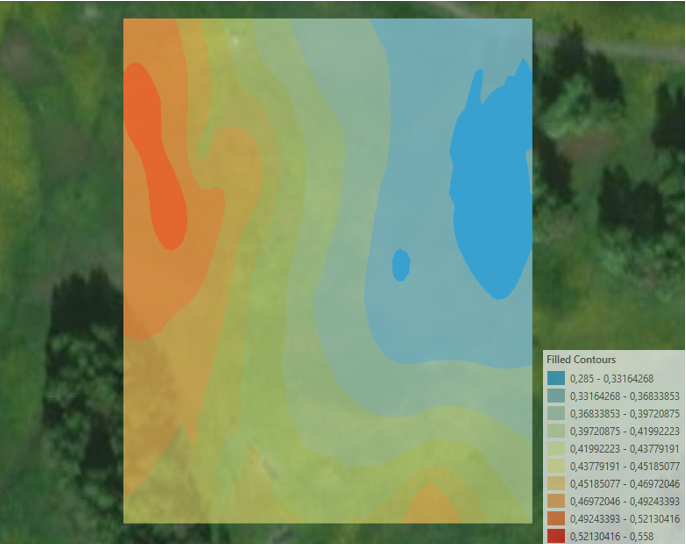}
    \caption{Spatial variation of soil moisture over a grass-covered field in Luleå, Sweden. These spatial variations result from Ordinary Kriging interpolation, and the gradient colours show soil moisture variation.}
    \label{fig:SM_variation}
\end{figure}

Soil Moisture (SM) is a crucial variable in the hydrological cycle, and its knowledge is vital for flood and drought forecasting. 
Furthermore, soil moisture parameter influences agriculture and water-resources management; therefore, accurate measurement and prediction of SM are of great importance \cite{lamichhane2025soil}, \cite{mondal2024quantifying}, \cite{tramblay2022estimating}. 
It changes dynamically in space and time and has high spatial heterogeneity in SM fields, as shown in Fig.\ref{fig:SM_variation}, even within small catchments \cite{zhou2007temporal},\cite{liu2025spatiotemporal}. 
Knowing the spatio-temporal variability is important for realising and modelling the hydrological processes \cite{pandey2010spatial}.
Among spatial and temporal variability, a correlation has been examined, revealing that the temporal stability of SM is associated with the temporal persistence of its spatial distribution patterns \cite{martinez2003temporal}.

SM dynamics are affected by nonlinear interactions with environmental and physical factors, such as climate, soil, vegetation and topography\cite{alahmad2025spatiotemporal}, \cite{liu2025spatiotemporal},\cite{singh2023strategies}.
Recent work by \cite{settu2025data} presents that Artificial Neural Networks (ANNs) and tree ensembles such as Random Forests (RFs) outperform linear models as they capture complex nonlinear interactions among hydrotographic and climate predictors. 
Furthermore, the nonlinear dependence of SM on multiple features is emphasised when RFs and ANNs significantly improve the estimation compared to simpler relationships \cite{ondieki2023enhancing}, \cite{alahmad2025spatiotemporal}.

Another challenging issue is the mismatch between spatial scales and the multisource assimilation.
The study by \cite{vergopolan2021smap} directly addresses the spatial scale mismatch between satellite and field scales by using inverse Hydroblocks-RTM to downscale SMAP to 30m. 
Furthermore, data assimilation of in-situ and remote sensing SM improves the model states, as reported by \cite{kivi2023comprehensive}, but also highlights the limitations of remote sensing for the topsoil at approximately $~5cm$ depth and the need for downscaling and error modelling to address scale mismatches. 

Traditional physics-based process models describe the hydrological processes governing SM transfer using physical equations and compute the relevant explanatory variables as part of land-surface data assimilation techniques \cite{fu2023soil}.
These models are mostly numerical and rely heavily on highly accurate input data. 
In addition, these mechanistic models demand substantial computational costs, which restricts their broader application at large scales \cite{zhang2025evolution}
The statistical methods were then introduced to improve model adaptability.  
Researchers can utilise data-driven approaches to derive empirical relationships between SM and environmental parameters, thereby reducing reliance on physical parameters \cite{zhang2025evolution}, \cite{wang2023comprehensive}. 

The objective of this study is to survey data-driven models for SM regression and classification tasks and to compare or distinguish between them. 
The focus will be on categorising the models by learning paradigm, and on highlighting current challenges and research gaps.

\section{Soil Moisture Modelling Tasks}
\label{sec:tasks}

\begin{figure*}[t]
    \centering
    \includegraphics[width=\textwidth]{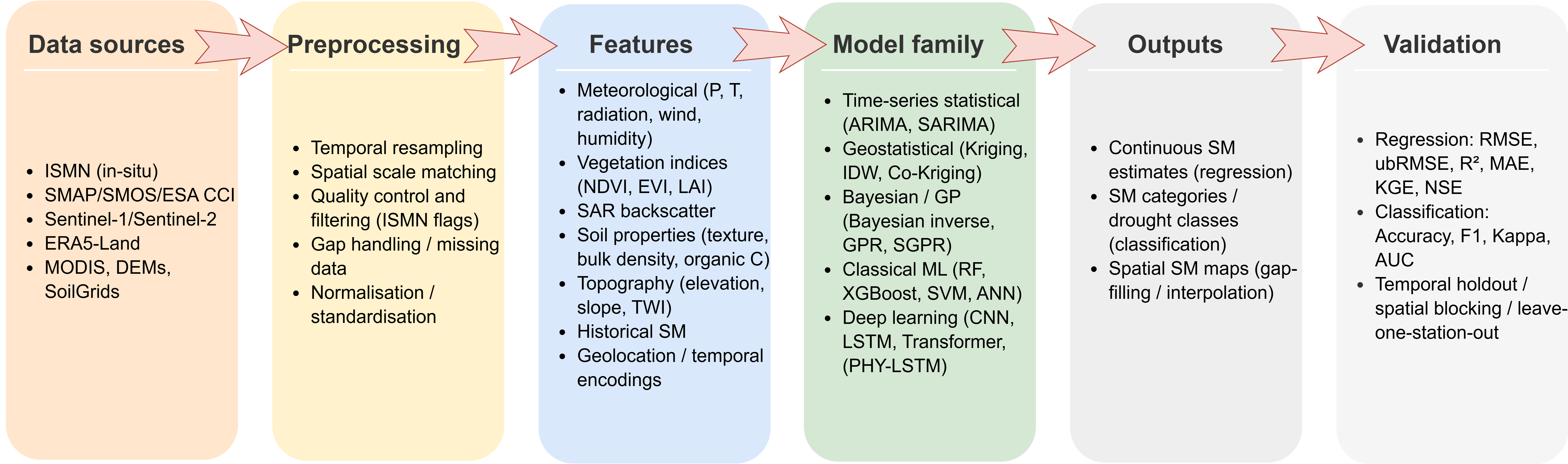}
    \caption{Pipeline for data-driven soil moisture modelling, from data acquisition and preprocessing to feature construction, model selection, outputs, and validation. }
    \label{fig:workflow_sm}
\end{figure*}

Data-driven SM modelling is commonly formulated as either a regression or a classification problem, depending on the target variable and application context. 
Regression predicts continuous SM values, while classification assigns samples to discrete moisture states such as dry, moderate, or wet \cite{wang2023comprehensive}, \cite{tunccay2023application}, \cite{zhang2025hyperspectral}. 
In addition, some studies address spatial downscaling, where coarse-resolution SM products are refined using higher-resolution auxiliary data\cite{vergopolan2021smap}, \cite{senanayake2024}, which fall under the umbrella of regression tasks.
Despite their inherent differences, the assumptions made and the way they utilise the provided input-output data all follow the same step shown in Fig.~\ref{fig:workflow_sm}.

\subsection{Input Variables and Data Sources}
\label{sec:inputs}

In all data-driven approaches, the GIGO aphorism holds (Garbage In Garbage Out). Therefore, of paramount importance in any of those tasks is the appropriate selection of input variables. 
In the field of SM of as iputs to the modelling tasks are meteorological, remote sensing, soil, topographic, and temporal predictors and combinations of them including, but not limited to precipitation, temperature, solar radiation, vegetation indices, SAR backscatter, soil texture, elevation, slope, and antecedent soil moisture \cite{alahmad2025spatiotemporal}, \cite{liu2025spatiotemporal}, \cite{nguyen2022low}, \cite{ondieki2023enhancing}, \cite{navidi2022predicting}. 
In-situ measurements from the International Soil Moisture Network (ISMN), passive microwave products (e.g. SMAP and SMOS), Sentinel-1 SAR observations, ERA5-Land reanalysis, and auxiliary products (e.g.  MODIS, DEMs, and SoilGrids) are among the most common data sources \cite{liu2025smrfr}, \cite{tong2021spatial}, \cite{vergopolan2021smap}, \cite{vahidi2025multi}. 

\subsection{Regression Tasks}
\label{sec:regression}

Regression is the most commonly employed tool in SM research, given the fact that SM measurements "live" in a continuous space. 
To put it formally, the regression paradigm tries to estimate or forecast moisture given a set of environmental and geospatial predictors, either static or in a chronologically ordered setting \cite{wang2023comprehensive}. 
This way, the problem is formulated as a standard supervised setting, trying to create a mapping between a group of predictors and a moisture value, either recorded on the surface or underground.

Regression can be applied into different settings depending on the input and more importantly the output space: a) \textit{point estimation}, with the goal of predicting SM at a specific location or depth from contemporaneous predictors \cite{wang2023comprehensive}, \cite{liu2025smrfr} b)  \textit{temporal forecasting}, with the goal of predicting SM values in the future using historical observations and exogenous inputs (e.g. precipitation, temperature, evapotranspiration, etc) \cite{adhikari2013introductory}, \cite{box2015time}, \cite{giorgio2022soil} c)  \textit{multi-depth prediction}, which differs from the first category that models are trying to predict SM values across several soil layers \cite{alahmad2025spatiotemporal}, \cite{vahidi2025multi} and d) \textit{gap-filling} or \textit{record reconstruction}, where the goal is to "impute" SM observations in sparse or incomplete datasets \cite{zhang2021reconstruction}, \cite{tong2021spatial}.Last but not least, a special case of regression modeling, which deserves mentioning separately, is \textit{spatial downscaling} where SM measurements are refined to finer spatial resolutions using more densely located auxiliary variables (e.g vegetation maps, land surface temperature, topography, etc.) \cite{vergopolan2021smap}, \cite{senanayake2024}, \cite{soltani2024}. Although downscaling is sometimes discussed separately because of its practical importance, it is best interpreted as a specialised regression problem in which the target remains continuous SM but the emphasis shifts to recovering finer spatial detail.

\subsection{Classification Tasks}
\label{sec:classification}

Classification modelling relaxes the need to predict a continues variable by the task of categorizing SM to predefined categories/classes \cite{salakpi2022dynamic}, \cite{tramblay2022estimating}, \cite{ismail2025cnn}, \cite{zhang2025hyperspectral}. 
The class formation can either rely on expert knowledge or on binning continuous values \cite{tunccay2023application}. 
Even though this lack of precision may look at first glance strange, it can be useful in case not enough data are available in order to formulate a regression problem and/or in case the said categories are enough as part of a broader system or because they are easier to interpret (e.g. for applications such as  
 drought monitoring, irrigation decision support, to name a few).

\subsection{Evaluation and Validation}
\label{sec:evaluation}

As in every Data-Driven model, the utility of a model is evaluated based on a specific metrics that measure the output(s) of the model against a set of target values (e.g. $R^2$, RMSE, MAE, ubRMSE, KGE, etc., for regression and accuracy, F1-score, kappa, precision, recall,  AUC, etc., for classification \cite{wang2024comprehensivedl}, \cite{salakpi2022dynamic}. 
The way that metric is estimated (design of training/validation/testing sets) is also critical. Random cross-validation should probably be avoided in favour of temporal holdout, spatial blocking, or leave-one-station-out validation \cite{liu2025smrfr}, \cite{roberts2017}. 
The overall design of the experimental set-up is crucial in order for the proposed modelling approach to generalise to new/unseen data.

\section{Data-Driven Model Taxonomy}
\begin{figure*}
    \centering
    \includegraphics[width=\textwidth]{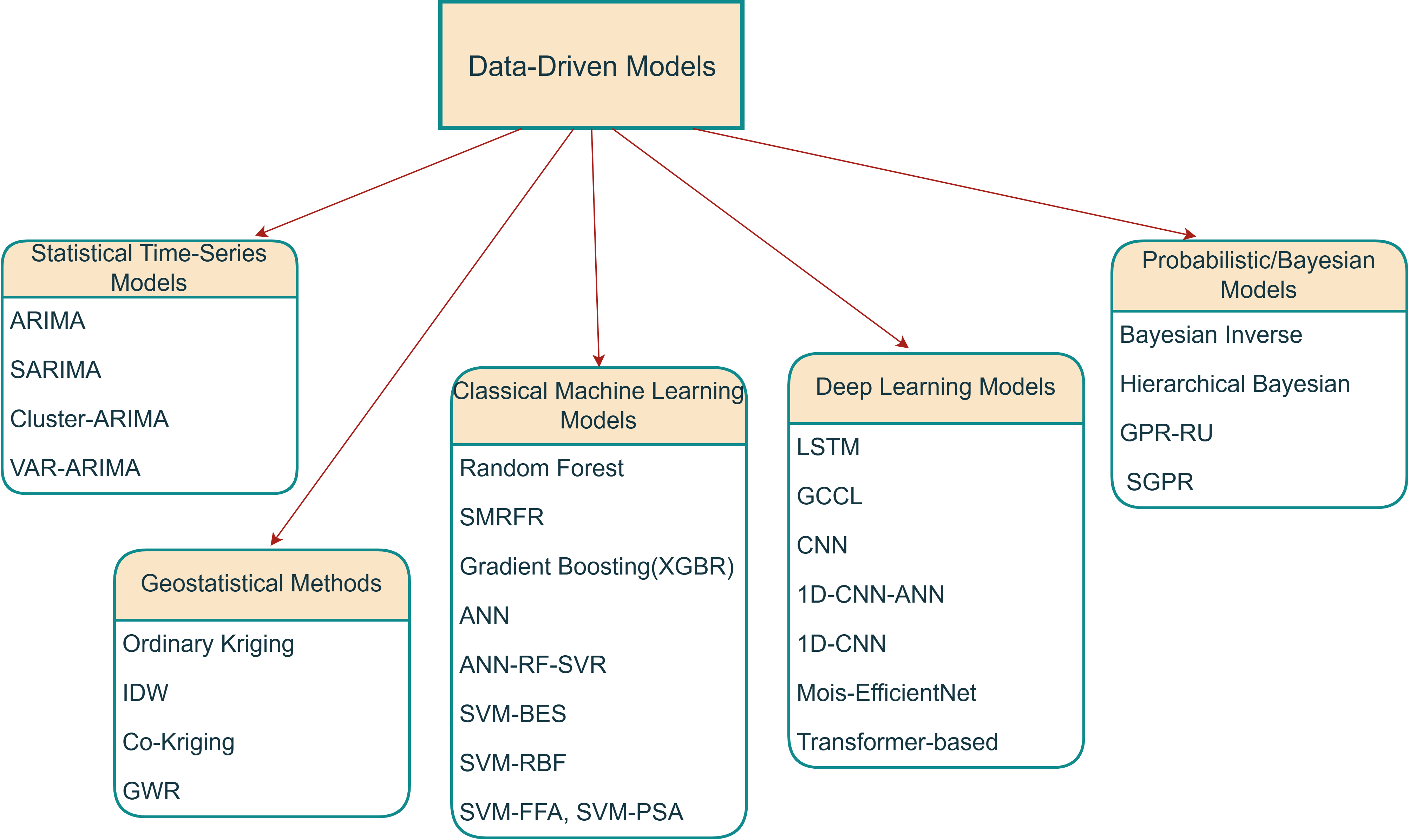}
    \caption{Diagram with the categorisation of this survey's data-driven models}
    \label{Data_driven_models}
\end{figure*}

This section reviews representative data-driven approaches for SM regression and classification, starting with the more established ones and moving to more recent trends. 
The taxonomy of survey's data-driven models is summarised in Fig.\ref{Data_driven_models}.

\subsection{Statistical Time-Series Models}
A statistical time-series model is a data-driven model that assumes the observed data are generated by a random process and describes the relationship between dependent and independent variables, solving an optimisation problem \cite{davison2003statistical}. 
The main advantages of statistical modelling include interpretability, robustness under limited data availability, and the ability to provide confidence intervals for predictions, which however, comes at a cost of the need to make some assumptions regarding the distributions of the data and the generative process \cite{storm2024probabilistic}, \cite{volodina2021importance}, \cite{hagar2025efficient}. Most of them have been around for over half a century. 

Autoregressive Integrated Moving Average (ARIMA) model uses past observation errors as inputs in order to forecast future values. Variations of the basic model have found extensive use in SM modelling. Huang in \cite{huang2023research} used a plain ARIMA (12,1,0) for predicting the precipitation and the soil evaporation. An extension of the basic technique is the Seasonal ARIMA (SARIMA), designed to handle seasonal time series \cite{adhikari2013introductory}. In \cite{box2015time} the SARIMA model utilises historical SM time-series at a specific depth to predict SM while in \cite{balasooriya2022forecasting}, the SARIMA Model is used to forecast the moisture content in precision agriculture. 
Giorgio et al. in \cite{giorgio2022soil} developed a SARIMA-based time-series data-driven model for a 48-h forecast of soil water content and salinity in the context of irrigation with reclaimed water, using data from soil water content and salinity data from 50 cm beneath the soil surface with a time resolution of 15 min, hourly atmospheric data and daily irrigation amounts.
In \cite{wang2024forecasting}, the Cluster-ARIMA model is used to predict soil respiration wherein the soil respiration data is partitioned into clusters using clustering techniques and then  Seasonal ARIMA (SARIMA) is employed to determine the cluster to which the data to be forecasted belongs. The prediction accuracy from the experimental results was 98,3\%.
The Vector Autoregression VAR-ARIMA model was used by \cite{wenresearch} to predict SM, treating precipitation, SM, and evapotranspiration as independent inputs.

\subsection{Geostatistical Methods}
Geostatistical models are spatial modeling approaches that consider SM as a realisation of a spatial stochastic process  \cite{diggle2003introduction}, \cite{cressie2015statistics}. 
Unlike classical regression models that assume independent observations, geostatistical methods model the dependence between measurements as a function of spatial distance \cite{cressie2015statistics}.
Kriging is the most widely used spatial predictor, and it outputs spatial interpolation of the collected data and generates a digital soil map of the variation in the predicted variable \cite{hilal2024geostatistical}.
In \cite{tong2021spatial}, the Ordinary Kriging (OK), utilised the available Soil Moisture Active Passive (SMAP) L3 SM pixels for filling the gaps and interpolating complete SM data. 
The cross-validation results showed a high correlation with the official SMAP SM and a high coefficient of determination.

In \cite{rash2024spatial}, researchers utilised Inverse Distance Weighting (IDW) and Geographic Weighted Regression (GWR) for estimating the spatial distribution of soil properties and compaction characteristics. 
At the same time, the independent variable was the Modified Normalised Difference Water Index (MNDWI) from the Landsat8 satellite image.
An assessment of OK and Ordinary Co-Kriging (OCK) for spatial interpolation of precipitation by month and quarter is presented in \cite{usowicz2021improvement}.
The input data are precipitation data collected over 4-5 years from rain gauges, and for the OCK, they utilised spatiotemporal data of SM from the SM and Ocean Salinity (SMOS) global satellite, in addition to precipitation data as an ancillary variable.
OCK showed the highest predictive performance for spatial precipitation distribution in both quarterly ($R^2$=0.944–0.992) and monthly analyses, clearly outperforming OK and IDW, which exhibited substantially lower $R^2$ values.

\subsection{Classical Machine Learning Models}
ML models automatically learn functional relationships between input variables and outputs by optimizing model parameters from input-output pairs without the need to exploit expert knowledge \cite{zhou2021machine}.

RFs is a purely data-driven ML model (belonging in the family of ensemble methods) that can capture nonlinear relationships and perform well on heterogeneous data \cite{breiman2001random}. 
For example, in \cite{liu2025smrfr} the SMRFR (Soil Moisture RF Regression) model took as inputs the combination of several SM variables, sucsh as in-situ SM data from ISMN (International Soil Moisture Network), ERA5-Land reanalysis multi-source predictors, MODIS vegetation indices, soil characteristics and topographic attributes to estimate SM at five depth layers daily and globally with 9 Km spatial resolution for the period 2000-2023.
In \cite{tramblay2022estimating}, RFs were utilised for precipitation, land use, altitude, and potential evapotranspiration to estimate Soil Water Holding Capacity. 
Additionally, the RFs approach was compared with direct estimation using pedotransfer functions from soil maps.
The results showed the RFs approach was more robust, especially for low soil moisture values.
Another study \cite{carranza2021root} examined RFs for interpolating and extrapolating Root Zone Soil Moisture (RZSM) estimates at the daily timescale. 
The input comprised in-situ measurements from an agricultural catchment collected over two years.
The results after the comparison of RFs prediction for RZSM with simulation from process-based models showed higher accuracy for the RFs.

For high-resolution SM prediction, in \cite{nguyen2022low}, the Extreme Gradient Boosting Regression (XGBR) (another member of the family of ensemble methods) was tested in combination with a Generic Algorithm (GA), using data fusion from Sentinel-1, Sentinel-2, and ALOS Digital Surface Model (DSM).
The result showed that XGBR-GA achieved the highest performance among RFs, Support Vector Machines, and CatBoost GB regression, with $R^2$=0.891 and RMSE=0.879\%.
An ANN was proposed in \cite{zhang2021reconstruction} for reconstructing missing daily surface SM records from the ESA CCI.
The considered variables were soil texture, geographic and topographic features, and vegetation conditions, among others.
The ANN model results outperformed the results of an OK model especially in regions with sparse vegetation.
Another study \cite{lee2022improved} evaluated an ANN model combined with other ML-based models, such as RFs and Support Vector Regression (SVR), for daily improved SM data production. 
Data from satellites and the Land Surface Model (LSM) were utilised, and the evaluation using the International Soil Moisture Network (ISMN) showed the ML-based ensemble's robustness in complex topographically areas with high-density vegetation.

Support Vector Machines (SVMs) are kernel-based supervised learning models used for both regression and classification and are attractive because they can perform well under limited or noisy data.
In \cite{huang2023improved}, the SVM model was enhanced using the Bald Eagle Search (BES) algorithm for SM prediction outperforming conventional SVMs. 
Navidi et al. \cite{navidi2022predicting} compared ANNs, Adaptive Neuro-Fuzzy Inference System (ANFIS), SVMs and an optimised SVM by firefly and particle swarm meta-heuristic algorithms, named SVM-FFA and SVM-PSA for predicting soil water content, using geometric mean diameter of soil particles, bulk density, organic carbon, NDVI and NSMI, with the metaheuristically optimised methods outperforming the conventional ML approaches. 
In \cite{mumtaz2025utilising}, an SVM, an RF, a GB, and an LR model were tested for predicting drought events using average temperature, specific humidity, soil moisture and dew point.
The SVM model outperformed the other ML models, achieving an AUC of 0.8166.

\subsection{Deep Learning Models}
DL models are methods based on deep (multilayer) NNs (i.e. stacking nonlinear layers together). What makes them attractive is their ability to learn hierarchical feature representations directly from raw data, eliminating in most cases the need for handcrafted features \cite{amanullah2020deep}.
In \cite{celik2022soil} a Long-Short-Term-Memory (LSTM) model acting on satellite data, climate time series, topographic and soil-type data forecasted daily SM values.
achieving ($R^2$) equal to 0.87 and RMSE equal to 0.046.
In \cite{pan2025combining}, a combination of Graph Convolutional LSTM (GConvLSTM) and Convolutional LSTM (ConvLSTM), named GCCL utilised historical time series coming from SMAP L4 SM spatiotemporal grids for forecasting SM for the next 7 days.
achieving RMSE values between 0.018–0.038 m³/m³ for 1–7 day forecasts and reduced error by up to 14.3\% compared to ConvLSTM. 

A Convolutional Neural Network (CNN), another example of ANNs, was used to forecast SM levels, as part of an AI irrigation framework \cite{ismail2025cnn}, with the results ($R^2$=0.87\%) and significantly exceeding those of linear regression and other baseline models.
In \cite{vahidi2025multi}, the 1D-CNN-ANN model is presented to estimate SM in different depths by utilising raw amplitude data from Ground Penetrating Radar (GPR). 
The achieved performance in terms of $R^2$ and RMSE outperformed those of GBMs and SVMs in 10-20-30 cm depth. The proposed model was robust even at 40 cm depth prediction.
For the classification of SM at 6-24inches depth, in \cite{zhang2025soil}, a 1D-CNN was developed, which utilised UGV-acquired hyperspectral data and spectral preprocessing methods. 
The results showed a testing accuracy of 0.67 and a strong discriminative ability (AUC) of 0.85.
The Mois-EfficientNet is proposed in \cite{liu2026mois} for classification and regression of root-zone soil moisture using GPR data-derived images. Initially, the Mois-EfficientNet classifies GPR images of tree roots into distinct moisture-content levels, and then this trained classification network is used to predict soil moisture. 

With the emergence of Transformer architecture, their use in SM modelling showed that they can be utilised for long prediction horizons and deep-soil estimation tasks \cite{wang2024transformer}, \cite{wang2024comprehensivedl}. In \cite{wang2024comprehensivedl}, where seven DL models were tested, Transformers, even though they did not outperform LSMTs, were competitive, especially for longer-horizon and deeper-soil prediction. However, in Wang and Zha \cite{wang2024transformer}, the Transformer model outperformed LSTM on average across all time lags, whereas LSTM-Transformer outperformed both for longer time lags. 
One of the latest trends in data-driven methods is physics-guided DL, where physically meaningful constraints are incorporated into the training objective. Geng et al. \cite{geng2024physically} proposed physically-guided LSTM models (PHY-LSTM) for next-day soil moisture forecasting using the global ERA5-Land dataset, and reported that PHYs-LSTM increased  $R^2$ by 20.7\%, and reduced RMSE by 8.2\%, compared with a conventional LSTM baseline.

\subsection{Probabilistic/Bayesian Models}
Probabilistic and Bayes models have been around for quite sometime. However recently with the advances in computational power have become relevant for data intensive applications.
A Bayesian inverse model was presented by \cite{somarathna2021mapping}, which utilised sparse data for predicting soil Available Water Capacity (AWC). 
The input was measurements of upper and lower drained limits across the tested area, and the output was the spatial prediction of AWC on a 90-m grid.
Salakpi et al. 2022 \cite{salakpi2022dynamic} compared a hierarchical Bayesian model with a regular Bayesian autoregression model to improve the accuracy and precision of agricultural drought forecasting across different regions.
The Hierarchical Bayesian model forecasted more accurate and precise drought probabilities and had a lower probability of false alarms.
A prediction method based on Gaussian Process Regression (GPR), incorporating the Radially Uniform (RU) design algorithm to reduce computational cost, was proposed by \cite{liu2020short}.
The experimental results show that the GPR model with the RU design algorithm outperforms the generic GPR model, achieving lower deterministic and probabilistic forecasting errors and reduced training time.
Zhang et al. \cite{xue2022ensemble} introduced Stacking GPR (SGPR), an ensemble-learning-based method for soil moisture estimation, which notably exceeded existing models.

\section{Discussion - Conclusions}
\begin{table*}[htbp]
\caption{Summary of Data-Driven Models for Soil Moisture}
\label{tab:models}
\centering
\scriptsize 
\begin{tabularx}{\textwidth}{p{2.0cm} X p{4.0cm}}
\toprule
Model & Input & Output \\
\midrule
ARIMA\cite{adhikari2013introductory} & Past observation errors & Forecast future values\hfill (R)\\
Cluster-ARIMA\cite{wang2024forecasting} & Clusters of past observations & Predict soil respiration \hfill (R)\\
VAR-ARIMA\cite{wenresearch} & Precipitation, Soil Moisture, Evapotranspiration & Predict Soil Moisture\hfill (R)\\
SARIMA\cite{adhikari2013introductory} & Historical SM data at specific depth & Forecast Soil Moisture at the same depth\hfill (R)\\
OK\cite{tong2021spatial} & SMAP L3 Soil Moisture pixels & Complete SM data for filling gaps and interpolation\hfill (R)\\
IDW-GWR\cite{rash2024spatial} & MNDWI from Landsat 8 satellite image & Estimation of spatial distribution of SM\hfill (R)\\
OCK\cite{usowicz2021improvement} & Precipitation and spatiotemporal SM data from SMOS & Spatial interpolation of precipitation by month and quarter\hfill (R)\\
SMRFR\cite{liu2025smrfr} & In-situ SM, MODIS vegetation indices, soil characteristics and topographic attributes & Estimation of SM at 5 depths daily with 9km spatial resolution\hfill (R)\\
RF\cite{tramblay2022estimating} & Temperature, precipitation, land use, altitude, evapotranspiration & Estimation of Soil Water Holding Capacity\hfill (R)\\
RF\cite{carranza2021root} & In-situ measurements from a catchment over 2 years & Estimation of Root Zone Soil Moisture daily\hfill (R)\\
XGBR - GA\cite{nguyen2022low} & Data fusion from Sentinel-1,2 and ALOS Digital Surface Model & High resolution SM prediction\hfill (R)\\
ANN\cite{zhang2021reconstruction} & Geographic and topographic features, soil texture and vegetation conditions & Reconstructing missing daily SM records from ESA CCI\hfill (R)\\
ANN-RF-SVR\cite{lee2022improved} & Data from satellites and LSM & Daily improved SM data production\hfill (R)\\
SVM-BES\cite{huang2023improved} & BES algorithm & Improves the accuracy and efficiency of SM prediction\hfill (R)\\
SVM-FFA, SVM-PSA\cite{navidi2022predicting} & Geometric mean diameter of soil particles, bulk density, organic carbon, NDVI, NSMI & Predict SM\hfill (R)\\
SVM-RBF\cite{mumtaz2025utilising} & Average temperature, specific humidity, soil moisture and dew point & Drought occurrence\hfill (C)\\
LSTM\cite{celik2022soil} & Radar satellite, climate time series, topographic and soil-type data & Forecast daily SM values\hfill (R)\\
GCCL\cite{pan2025combining} & Historical SM time series from SMAP L4 SM spatiotemporal grids & Forecast SM for the next 7 days\hfill (R)\\
CNN\cite{ismail2025cnn} & Environmental indicators & Forecast SM levels\hfill (R)\\
1D-CNN-ANN\cite{vahidi2025multi} & Raw amplitude data from Ground Penetrating Radar & Estimation of SM in different depths\hfill (R)\\
1D-CNN\cite{zhang2025soil} & UGV-based hyperspectral data(500-1000nm) and spectral preprocessing methods & Soil moisture class at6-24inches depth\hfill (C)\\
Mois-EfficientNet\cite{liu2026mois} & GPR data-derived images & Distinct levels of root moisture content and prediction of SM\hfill (C),(R)\\
Transformer-based\cite{wang2024transformer} & Groundwater level and meteorological data & Forecast SM for longer prediction horizons and deeper soil estimation\hfill (R)\\
PHY-LSTM\cite{geng2024physically} & ERA5-Land Dataset & Forecast next-day SM\hfill (R)\\
Bayesian Inverse\cite{somarathna2021mapping} & Upper and Lower drained limits & Spatial prediction of Available Water Capacity\hfill (R)\\
Hierarchical Bayesian\cite{salakpi2022dynamic} & Historical vegetation condition data and spatial grouping variables & Forest accurate and precise drought probabilities\hfill (C)\\
GPR-RU\cite{liu2020short} & Historical SM time series, lagged SM values, RU selected representative training samples & SM point forecast, Predictive variance and intervals\hfill (R)\\
SGPR\cite{xue2022ensemble} & 11 multisource remote sensing features & Estimation of SM\hfill (R)\\
\bottomrule
\end{tabularx}
\vspace{1mm}
\footnotesize{\textit{R}: Regression task; \textit{C}: Classification task.}
\end{table*}

The reviewed literature, summarised in Table \ref{tab:models}, shows that several data-driven techniques have been used for SM modelling, with the input variables and the predicted output for each approach. 
Statistical and geostatistical approaches remain relevant; however, ML and DL are gaining popularity due to their ability to deal with heterogeneous and nonlinear inputs. 
The main issue in comparing the proposed methods is the variability in experimental design (e.g., various inputs and outputs).  

The majority of studies focus on Regression because SM is inherently continuous, but there are still a few studies that utilise classification. 
Future work should place greater emphasis on uncertainty quantification and benchmarking across regions, depths, and sensing modalities. 

Overall, data-driven methods offer strong predictive capability for complex SM regression and classification problems. However, their effectiveness is often limited by data availability and reduced interpretability.

\bibliographystyle{splncs04}
\bibliography{bibliography}

\end{document}